\documentclass{article}
\usepackage{pgfplots}
\usepackage{spconf,amsmath,graphicx}
\usepackage{booktabs}
\usepackage[english]{babel}
\usepackage[utf8]{inputenc}
\usepackage{subcaption}

\usepackage{algorithm}
\usepackage[noend]{algpseudocode}
\algrenewcommand\algorithmicrequire{\textbf{Input:}}
\algrenewcommand\algorithmicensure{\textbf{Output:}}
\title{Automated Audio Captioning with Epochal Difficult Captions for curriculum learning}

\name{Andrew Koh{*}, Soham Tiwari{'}, Chng Eng Siong{*}}
\address{Nanyang Technological University{*} \thanks{This research is supported by ST Engineering Mission Software \& Services Pte. Ltd under collaboration programme (Research Collaboration No: REQ0149132).}, Manipal Institute of Technology{'}}
\pgfplotsset{compat=1.17}
\tikzstyle{line}=[draw] 
\begin{document}
\maketitle
\begin{abstract}
In this paper, we propose an algorithm, Epochal Difficult Captions, to supplement the training of any model for the Automated Audio Captioning task. Epochal Difficult Captions is an elegant evolution to the keyword estimation task that previous work have used to train the encoder of the AAC model. Epochal Difficult Captions modifies the target captions based on a curriculum and a difficulty level determined as a function of current epoch. Epochal Difficult Captions can be used with any model architecture and is a lightweight function that does not increase training time. We test our results on three systems and show that using Epochal Difficult Captions consistently improves performance.

\end{abstract}
%
% \begin{keywords}
% One, two, three, four, five
% \end{keywords}
%
\section{Introduction}
\label{sec:intro}

The Automated Audio Captioning task is a cross-modal task where the model is trained to generate a descriptive sentence for an input audio. This task has many practical real world use cases, such as closed captioning for people who are deaf or hard of hearing. Automated Audio Captioning describes the sound events that are happening in both the foreground and background of the audio, as opposed to Automated Speech Recognition, which predicts the low time frame phonemes and utterances.

The Automated Audio Captioning task requires the model to be able to capture both high time frame and low time frame events that are occurring throughout the audio. As such, previous work have often used pretrained models such as PANNs that have already been trained on audio tagging in order to aid the model in generating factually correct caption. In a similar vein, several authors \cite{koizumi_transformer-based_2020, zhonjie2021integrating} have used keyword estimation, a supplementary and intermediate objective where the model has to predict one or a few keywords from the encoder to help the decoder to generate better captions. The keyword estimation task has shown to have potential in helping the model perform better. However, the keyword estimation task requires heavy preprocessing to create the audio-keywords pairs for training. Furthermore, while the different authors have the same motivation behind keyword estimation task, their implementations and audio-keywords pairs are often different and subject to their own interpretation. This makes it hard to compare results and methods across different papers. We find that there is an elegant solution to this. Epochal Difficult Captions modify the training targets directly during the training loop. Based on a curriculum, the training captions start off with a higher keywords to sentence ratio during the earlier epochs in the the training run. This is done by removing stopwords directly from the batched training targets. As the training progresses, the keywords to sentence ratio will be gradually reduced until the sentence reverts to its original unmodified sentence.

We test Epochal Difficult Captions using two different publicly available systems \cite{koh2021automated, Mei2021} from previous work. Both systems are of the typical encoder-decoder structure. The encoder consists of either a single \cite{Mei2021} pretrained CNN or both \cite{koh2021automated} a pretrained CNN and transformer encoder. The decoder used is a 2 layer transformer decoder. The extracted audio features from the encoder is passed to the decoder for cross attention. Both systems are trained using cross entropy, but \cite{Mei2021} has the additional option of using reinforcement learning to optimize the metrics directly. We experiment using both cross entropy, and reinforcement learning where available. We find that using Epochal Difficult Captions consistently improves performance across these settings with no noticeable extra compute time and power needed. Epochal Difficult Captions is used only during training and hence does not affect any part of the inference process.

The rest of this paper follows this structure. In section \ref{sec:related_work }, we introduce related work pertaining to Automated Audio Captioning datasets, popular model architectures and training objectives. We also discuss the keyword estimation task, a popular intermediate task in Automated Audio Captioning. In Section \ref{sec:epochal_difficult_captions}, we explain our approach, Epochal Difficult Captions for Curriculum Learning. This section shows how we derive our difficulty curve and levels, and how stopwords are removed as an function of the difficulty level. In Section \ref{sec:experimental_details}, we specify our experimental settings, details, and process to evaluate the effectiveness of Epochal Difficult Captions. In Section \ref{sec:experimental_results_analysis}, we examine the performance of the models trained with Epochal Difficult Captions for Curriculum Learning. Finally, we give our concluding remarks in Section \ref{sec:conclusion}.

\section{Related Work} \label{sec:related_work }
% Automated Audio Captioning has been tackled by various authors. 

\subsection{Audio Captioning Datasets} 
\label{audio_cap_dataset}
The Clotho Dataset v2.1 \cite{drossos_clotho_2019} consists of 6974 audio samples of duration from 15 to 30 seconds. Each audio clip has 5 corresponding captions annotated by humans. Each caption has a length of 8 to 20 words, and the captions describe the events in that audio. The Clotho Dataset v1 was augmented by including more training samples and a validation split to form Clotho v2.1. The Clotho Dataset was created to address the critiques of the larger Audiocaps dataset \cite{kim_audiocaps_2019}, which was found to contain biases in the annotation process. There are also other niche datasets like the Hospital Scene and Car Scene dataset \cite{xu_audio_2020} which are Mandarin annotated.

In this work, we focus only on the Clotho Dataset v2.1, henceforth referred to as the Clotho Dataset.

\subsection{Model Architectures and Objectives}
\label{model_arch_objectives}

Most work use an encoder-decoder model architecture. The encoder is used to encode audio information to produce audio features, which is then passed to the decoder for cross attention and to generate the caption. Most architectures converge to a similar theme - using a pretrained convolutional neural network such as PANNs for encoding, and a transformer decoder for decoding. However, the training process often differs from the pretraining stage and an intermediate or supplementary objective.

For instance, the AT-CNN \cite{xu:2021:ICASSP:02}, and the Audio Captioning Transformer \cite{mei2021audio} pretrains their model encoder on audio tagging as an pretraining task before finetuning their model directly on Automated Audio Captioning. Other authors also use self supervised supplementary objectives \cite{koh2021automated} or caption retrieval \cite{zhang2020bertscore} to guide the model training. Self-Critical Sequence
Training \cite{scst_rl}, a reinforcement learning tactic used in Image Captioning, has also been used to optimize the model for Audio Captioning. Keyword Estimation, which we will go into detail in Section \ref{subsec:keyword_estimation}, has also been a popular choice \cite{koizumi_audio_2020, zhonjie2021integrating} of an intermediate objective.

\begin{figure}[ht!]
    \centering
    \includegraphics[scale=0.03]{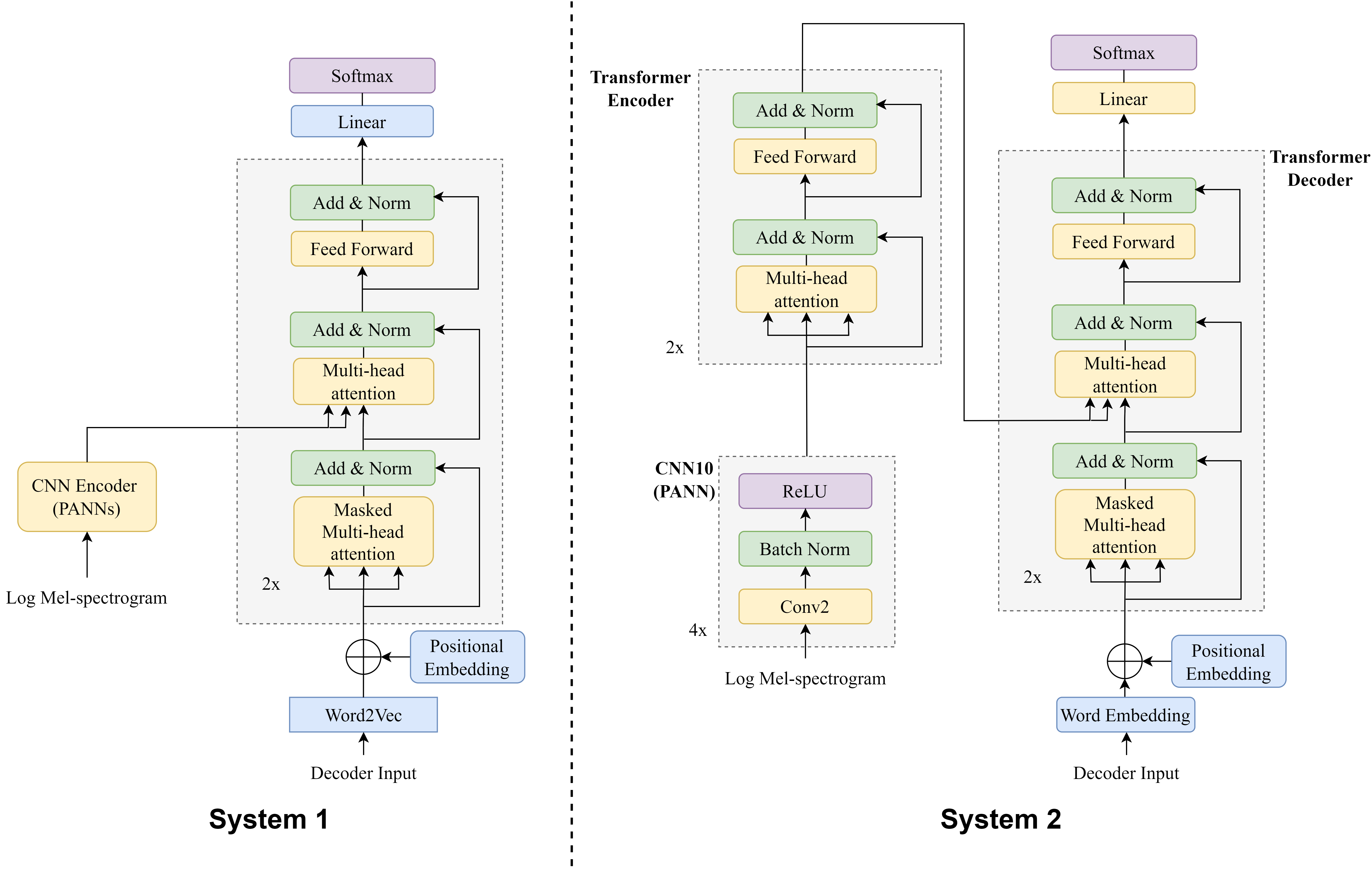}
    \caption{Detailed overview of System 1 and System 2. Both System 1 and 2 uses a pretrained CNN10 in the encoder. System 2 has an additional 2 layer transformer appended to the CNN10 in the encoder. For the decoder, both System 1 and 2 uses a transformer decoder. However, System 1 uses pretrained word2vec word embeddings, while System 2 trains the word embeddings from scratch.}
    \label{fig:system12}
\end{figure}

In this work, we build on the work of \cite{koh2021automated} and \cite{Mei2021} by using their model architectures that has been proven to achieve competitive results. Henceforth, we refer \cite{Mei2021} and \cite{koh2021automated} as System 1 and System 2 respectively. Figure \ref{fig:system12} delineates their system architectures.

% Several works has applied transfer learning to Automated Audio Captioning. Fine-tune PreCNN Transformer\cite{chen:2020:dcase:transformer_pretrained_cnn} used a 3 stage pipeline of first pretraining an encoder, then applying transfer learning to the pretrained model by freezing parts of the model for training, and finally finetuning the entire model. 

% AT-CNN \cite{xu:2021:ICASSP:02}, the current best performing model on Clotho, follows a similar approach with a 2 pipeline which first pretrains the CNN10 \cite{kong2020panns} / CRNN5 \cite{Dinkel_2020} encoder on an audio tagging task, and then performs transfer learning on the pretrained model on the audio tagging task. Similarly, the Audio Captioning Transformer \cite{mei2021audio} also pretrains the encoder on audio tagging before training on audio tagging.

% Other approaches uses specialized embeddings for transfer learning. \cite{koizumi_audio_2020} applied transfer learning to a pretrained GPT2 model, which was pretrained in the text modality. This approach also requires several other prerequisites, such as a BERTScore \cite{zhang2020bertscore} model, and several preprocessing steps for caption retrieval. In another vein, \cite{eren_audio_2021} used a combination of pretrained audio embeddings and subject-verb embeddings to generate captions. 

% In this work, we show that applying transfer learning using publicly available pretrained models in conjunction with the transformer encoder allows for significant performance improvements.

\subsection{Keyword Estimation} \label{subsec:keyword_estimation}

Several authors have applied the Keyword Estimation task as a intermediate task to aid the training of the model for Automated Audio Captioning. This is done by decomposing the Automated Audio Captioning task into two smaller tasks, Keyword Estimation and Caption Generation. The model encoder then performs Keyword Estimation to predict a set of keywords, which is then passed to the decoder for generation. The Keyword Estimation task helps to mitigate the problem of indeterminacy in word selection \cite{koizumi_audio_2020} hence reducing the solution space for the decoder to generate better captions.

Since the Clotho Dataset does not come with labels for keywords, authors often use their own interpretation to create labels for the keyword Estimation task. \cite{koizumi_audio_2020} uses a rule based keyword extraction method to produce keywords labels for training. Frequent word lemmas of nouns, verbs, adjectives, and adverbs as regarded as keywords. On the other hand, \cite{zhonjie2021integrating} extracts their keywords by using the Natural Language Toolkit\footnote{https://github.com/nltk/nltk} to discard `useless words' such as make, go, etc. Remaining verbs are used in their original forms, and the nouns are not changed. 

Firstly, given keyword estimation requires some form of pre-processing subject to the interpretation of the researcher, it is difficult to compare the effectiveness of the keyword estimation task across different pieces of work. Secondly, decomposing the Automated Audio Captioning task into subtasks keyword estimation and caption generation requires the decoder to rely on the prediction of the keywords from the encoder. This might result in error propagation if the keywords are inaccurate. With these considerations, we propose the Epochal Difficult Captions for curriculum learning, which has the same motivation behind keyword estimation, but does not require the decomposition of Automated Audio Captioning into subtasks.

% \subsection{Related Approaches} 

\section{Epochal Difficult Captions for Curriculum Learning} \label{sec:epochal_difficult_captions}
% describe what it is
Epochal Difficult Captions for Curriculum Learning is an algorithm to modify the target captions relative to the current stage of the training setup. The difficulty level, $D$, is a function of the current epoch. $D$ indicates how the curriculum modifies the target captions. In this paper, we use the stopwords curriculum. While we have another frequency-based curriculum, we omitted it due to space constraints.

\subsection{Difficulty level as a function of epoch}

The difficulty level in curriculum learning dictates the complexity of the captions and it should increase from a lower to a higher magnitude relative to the current epoch. We use an exponential function (Equation \ref{eq:diff}) followed by a floor and ceiling function (Equation \ref{eq:diff2}) to generate a difficulty value $D$ for all non-negative epoch values \(D  \in  (0, 1)\).

\begin{equation} \label{eq:diff}
    D^{\prime} = 1 - e^{( -\alpha * epoch )}
\end{equation}
\begin{equation} \label{eq:diff2}
  D=\begin{cases}
    0.05,           & \text{if } D < 0.05 \\
    D^{\prime},     & \text{otherwise}
  \end{cases}
\end{equation}

where the hyperparameter \(\alpha\) in Equation \ref{eq:diff} controls the rate of increase of difficulty. At $D = 1$, the original captions are not modified. The difficulty level should asymptotically tend to 1 nearing the max epoch. Therefore, \(\alpha\) requires the max epoch hyperparameter to be taken into consideration. In our experiments, we use \(\alpha = 0.20\) when max epoch is 30, \(\alpha = 0.10\) when max epoch is 60, and \(\alpha = 0.05\) when max epoch is 100.

% In our experiments, setting \(\alpha=0.05\) helped to achieve a difficulty value approximately equal to 1 at the \(100^{th}\) epoch. After the \(100^{th}\) epoch, the function generates values increasingly closer to 1

\begin{figure*}
\begin{subfigure}[b]{0.33\linewidth}
\begin{tikzpicture}[scale=0.62]
\begin{axis}[
    axis y line*=left,
    xlabel=$D$,
    ylabel=\text{total tokens per epoch},
    legend pos=south east
]
\addplot+[
    only marks,
    line,
    mark= *,
    mark size=1pt]
table[x=difficulty, y=total_tokens_per_epoch]
{100epoch.dat}; \label{plot_epoch_d_100}
\end{axis}
\begin{axis}[
    axis y line*=right,
    axis x line=none,
    ylabel=\text{average unique tokens per batch},
    legend pos=south east
]
\addplot+[
    only marks,
    color=red,
    line,
    mark= x,
    mark size=2pt]
table[x=difficulty, y=avg_tokens_per_batch]
{100epoch.dat};\label{plot_epoch_avgtokensperbatch_100}
\addlegendimage{refstyle=plot_epoch_d_100}\addlegendentry{\text{total tokens per epoch}}
\addlegendimage{refstyle=plot_epoch_avgtokensperbatch_100}\addlegendentry{average unique tokens per batch}
\end{axis}
\end{tikzpicture}
\caption{100 epochs} \label{plot:100_epoch_avgtokensperbatch}
\end{subfigure} 
\begin{subfigure}[b]{0.33\linewidth}
\begin{tikzpicture}[scale=0.62]
\begin{axis}[
    axis y line*=left,
    xlabel=$D$,
    ylabel=\text{total tokens per epoch},
    legend pos=south east
]
\addplot+[
    only marks,
    line,
    mark= *,
    mark size=1pt]
table[x=difficulty, y=total_tokens_per_epoch]
{60epoch.dat}; \label{plot_epoch_d_60}
\end{axis}

\begin{axis}[
    axis y line*=right,
    axis x line=none,
    ylabel=\text{average unique tokens per batch},
    legend pos=south east
]
\addplot+[
    only marks,
    color=red,
    line,
    mark= x,
    mark size=2pt]
table[x=difficulty, y=avg_tokens_per_batch]
{60epoch.dat};\label{plot_epoch_avgtokensperbatch_60}
\addlegendimage{refstyle=plot_epoch_d_60}\addlegendentry{\text{total tokens per epoch}}
\addlegendimage{refstyle=plot_epoch_avgtokensperbatch_60}\addlegendentry{average unique tokens per batch}
\end{axis}
\end{tikzpicture}
\caption{60 epochs} \label{plot:60_epoch_avgtokensperbatch}
\end{subfigure}
\begin{subfigure}[b]{0.33\linewidth} 
\begin{tikzpicture}[scale=0.62]
\begin{axis}[
    axis y line*=left,
    xlabel=$D$,
    ylabel=\text{total tokens per epoch},
    legend pos=south east
]
\addplot+[
    only marks,
    line,
    mark= *,
    mark size=1pt]
table[x=difficulty, y=total_tokens_per_epoch]
{30epoch.dat}; \label{plot_epoch_d_30}
\end{axis}
\begin{axis}[
    axis y line*=right,
    axis x line=none,
    ylabel=\text{average unique tokens per batch},
    legend pos=south east
]
\addplot+[
    only marks,
    color=red,
    line,
    mark= x,
    mark size=2pt]
table[x=difficulty, y=avg_tokens_per_batch]
{30epoch.dat};\label{plot_epoch_avgtokensperbatch_30}
\addlegendimage{refstyle=plot_epoch_d_30}\addlegendentry{\text{total tokens per epoch}}
\addlegendimage{refstyle=plot_epoch_avgtokensperbatch_30}\addlegendentry{average unique tokens per batch}
\end{axis}
\end{tikzpicture}
\caption{30 epochs} \label{plot:30_epoch_avgtokensperbatch}
\end{subfigure} 
\caption{Scatter plot of $D$ (x-axis) against total tokens per epoch (left y-axis) and average unique tokens per batch (right y-axis) for different epoch values 100 (a), 60(b), 30(c)} \label{fig:D_vs_unique_tokens_vs_total_tokens}
\end{figure*}
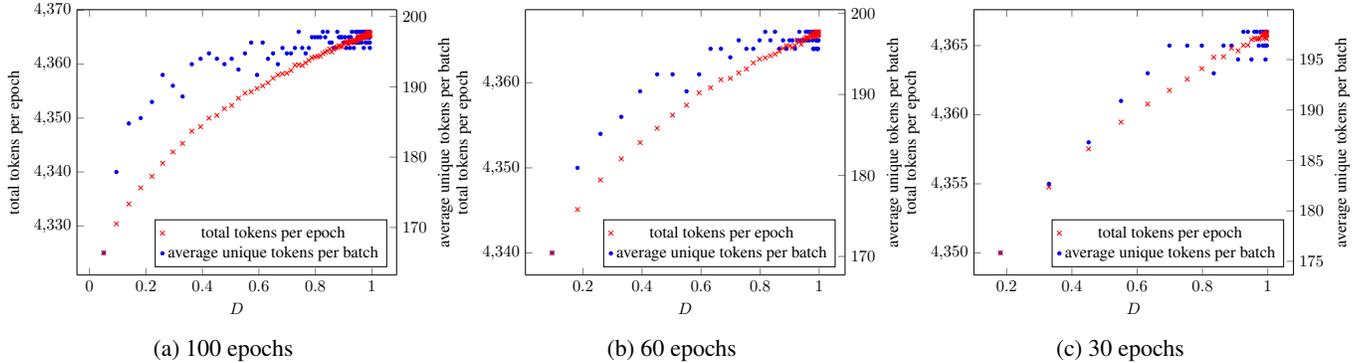

\subsection{Stopwords Curriculum}
In the stopwords curriculum, the target captions for each audio are modified by removing stopwords. Stopwords are common words that carry little useful information and can be obtained from the public nltk library. Some examples of stopwords are ‘for’, ‘do’, ‘its’, ‘yours’, ‘such’, ‘into’. Removal of these stopwords mimic the keyword estimation task described in Section \ref{subsec:keyword_estimation} by only including unique and less prevalent words in the caption.

\begin{algorithm}
    \caption{Stopwords Curriculum}
    \label{alg:stopwords}
    \begin{algorithmic}[1]
        \Require $y$ - the original target caption
        \Require $D$ - difficulty value (Equation \ref{eq:diff}, \ref{eq:diff2})
        \Require $SW$ - list of stopwords.
        \Ensure $y^\prime$ - modified caption.
        \State $P(remove) \gets (1 - D)$ 
        \State {$y^\prime \gets []$} 
        \For{$word \in y$} 
        \If{$word \in stop\_words$}
            \State{$s \gets$ sample uniform(0,1)}
            \If{$s <  P(remove)$}
                \State {$y^\prime \gets new\_sentence$}
            \Else
                \State {$y^\prime \gets y^\prime + word$}
            \EndIf
        \Else
                \State {$y^\prime \gets y^\prime + word$}
        \EndIf
        \EndFor
        \State \Return $y^\prime$
    \end{algorithmic}
\end{algorithm}

Algorithm \ref{alg:stopwords} describes the stopwords curriculum. Each stopword in the caption is an candidate for removal depending on the current difficulty level of the current epoch. The greater the difficulty level $D$ is, the lower the probability of removing a stopword from the caption. 

In Figure \ref{fig:D_vs_unique_tokens_vs_total_tokens}, we plot the relationship between $D$, the total number of words used per epoch in training, and the average number of unique words per batch during training. At low difficulty levels, the total number of tokens per epoch increases rapidly with each epoch at low difficulty levels. At higher difficulty levels above 0.9, the total number of tokens fluctuates around the same range. Using such a token distribution based on a fixed curriculum allows us to gradually expose the model to progressively more unique words.

\begin{table*}[!ht]
\centering
\resizebox{\textwidth}{!}{%
\begin{tabular}{@{}llllllllll@{}}
\toprule
Model                                    & BLEU\textsubscript{1} & BLEU\textsubscript{2} & BLEU\textsubscript{3} & BLEU\textsubscript{4} & ROUGE\textsubscript{L} & METEOR & CIDEr & SPICE & SPIDEr \\ \midrule

System 1 - cross entropy&	0.559&	0.358&	0.237&	0.153&	0.169&	0.374&	0.382&	0.116&	0.249\\
System 1 - cross entropy + stopwords&	0.558&	\textbf{0.362}&	\textbf{0.242}&	\textbf{0.159}&	\textbf{0.170}&	\textbf{0.375}&	\textbf{0.391}&	0.115&	\textbf{0.253}\\
\midrule
System 1 - scst&	0.641&	0.417&	0.277&	0.174&	0.182&	0.407&	0.432&	0.124&	0.278 \\
System 1 - scst + stopwords	&\textbf{0.642}&	0.409&	0.272&	0.172&	0.182&	\textbf{0.402}&	\textbf{0.444}&	0.124&	\textbf{0.284}\\
\midrule
System 2 - cross entropy & 0.553 & 0.367 & 0.248 & 0.160 & 0.162 & 0.372 & 0.359 & 0.111  & 0.235  \\ 
System 2 - cross entropy + stopwords    & \textbf{0.558}&	\textbf{0.376}&	\textbf{0.258}	& \textbf{0.172}&	\textbf{0.167}&	\textbf{0.376}&	\textbf{0.381}&	\textbf{0.115}&	\textbf{0.248} \\
\bottomrule
\end{tabular}
}

\caption{Comparison of performance of systems trained on Epochal Difficult Captions with the Stopwords curriculum against their counterpart}
\label{tab:results_stopwords}
\end{table*}

\section{Experimental Details} \label{sec:experimental_details}
\subsection{Data}
We use Clotho dataset v2.1 for all our experiments as mentioned in Section \ref{audio_cap_dataset}. Since we are building on the previous systems of two authors \cite{koh2021automated, Mei2021}, we use the same settings to preprocess the waveforms into the spectrograms as inputs into the models. For both systems, we use 64 Mel-bands, sampling rate of 44100, FFT window length of 1024, and a hop size of 512.

\subsection{Training and Evaluation}

For our experiments on both System 1 and System 2, we try to replicate the original work without using Epochal Difficult Captions to act as a control experiment, then rerun the same experiments with Epochal Difficult Captions. Where possible, we use the original hyperparameters.

For System 1 \cite{Mei2021}, we use a batch size of 32 with no gradient accumulation for 30 epochs and 60 epoch with no early stopping. System 1 has two different training setups. The first setup first optimizes on a cross entropy loss for 30 epochs, then uses a reinforcement learning technique, Self Critical Sequence Training (SCST) \cite{scst_rl}, to train the model for 60 epochs. Learning rate is set to $1 \times 10^{-4}$ for the cross entropy loss and $5 \times 10^{-5}$ for the reinforcement training. SpecAugmentation \cite{specaug} is applied to all log mel-spectrogram inputs as an data augmentation tactic. We do apply label smoothing. For inference, we perform beam search with beam size of 3 for decoding. 

For System 2 \cite{koh2021automated}, we use a batch size of 64 with a gradient accumulation steps of 4 for 100 epochs. Learning rate is set to $3 \times 10^{-4}$ and SpecAugmentation \cite{specaug} is applied to all log mel-spectrogram inputs as an data augmentation tactic. We do not apply label smoothing. For inference, we perform beam search with beam size of 4 for decoding.

The COCO image captioning evaluation process \cite{coco_dataset} is used to evaluate the generated caption. There are a few metrics used. The BLEU\textsubscript{n} scores \cite{bleu_cite} measures of n-gram overlap between the generated and reference caption, while ROUGE\textsubscript{L} \cite{lin-2004-rouge} bases its score off the longest common sequence. METEOR \cite{denkowski:lavie:meteor-wmt:2014} uses a harmonic mean of unigram precision and recall to score the generated sequence. CIDEr \cite{cider_cite} uses the average cosine similarity between the candidate sentence and the reference sentences to produce a score. Next, SPICE \cite{anderson2016spice} uses semantic propositional information to score the generated captions. Finally, the SPIDEr metric is a combination of both SPICE and CIDEr. We use all the aforementioned metrics for evaluation. The SPIDEr metric is however the metric that researchers try to beat.

% Our code is available at <link here>

\section{Experimental Results and Analysis} \label{sec:experimental_results_analysis}
We compare the effectiveness of using Epochal Difficult Captions for Curriculum Learning in 3 different settings. The results are shown in Table \ref{tab:results_stopwords}. From the table, we observe that the most meaningful metric, the SPIDEr score, increases across all 3 systems. For System 2 particularly, the effect of using Epochal Difficult Captions leads to an improvement of all metrics.

As described in Section \ref{sec:epochal_difficult_captions}, Epochal Difficult Captions allows the model to gradually transit from training on prominent keywords to the original unmodified caption. We believe this allows the model to focus on more important words such as nouns and action verbs in the earlier stage of the training, thus providing the model with a better ability to recognize objects and actions from the log mel spectrogram. As the training proceeds, filler words and syntactically important terms are consecutively added epoch-wise to enable the model to generate grammatically sound sentences.

\section{Conclusion} \label{sec:conclusion}
This work introduces the use of Epochal Difficult Captions for Curriculum Learning. Epochal Difficult Captions is an elegant evolution to the keyword estimation problem. We have examined the difficulty curve for Curriculum Learning and specified the stopwords Curriculum. We have shown that Epochal Difficult Captions can be easily added into any Automated Audio Captioning system during the training stage. Using Epochal Difficult Captions for Curriculum Learning, performance across 3 different settings in 2 systems improves across multiple metrics and the SPIDEr score.

\bibliographystyle{IEEEbib}
\bibliography{refs}

\end{document}